\documentclass{article} 
\usepackage{arxiv,times}
\iclrfinalcopy

\usepackage{amsmath,amsfonts,bm}

\def\eqref#1{equation~\ref{#1}}

\def\1{\bm{1}}

\DeclareMathAlphabet{\mathsfit}{\encodingdefault}{\sfdefault}{m}{sl}
\SetMathAlphabet{\mathsfit}{bold}{\encodingdefault}{\sfdefault}{bx}{n}

\usepackage{hyperref}
\usepackage{cleveref}
\usepackage{url}

\usepackage{graphicx}
\usepackage{bm}
\usepackage{wrapfig}
\usepackage{mathtools}
\usepackage{subcaption}
\usepackage{array}
\usepackage{booktabs}
\usepackage{multirow}

\title{FullPart: Generating each 3D Part at Full Resolution}

\author{
\\
  Lihe Ding$^{1}$\footnotemark[1], ~~Shaocong Dong$^{2}$\footnotemark[1], ~~Yaokun Li$^{1}$, ~~Chenjian Gao$^{1}$, ~~Xiao Chen$^{1}$, ~~Rui Han$^{3}$, \\
  Yihao Kuang$^{4}$, ~~Hong Zhang$^{4}$, ~~Bo Huang$^{4}$, ~~Zhanpeng Huang$^{3}$, ~~Zibin Wang$^{3}$, \\
  Dan Xu$^{2}$\footnotemark[2], ~~Tianfan Xue$^{1}$\footnotemark[2] \\
  \vspace{-2mm} \\
  {$^1$}CUHK 
  ~~~~{$^2$}HKUST 
  ~~~~{$^3$}SenseTime Research 
  ~~~~{$^4$}Chongqing University \\ \vspace{-2mm} \\
  {\tt\small\{dl023, tfxue\}@ie.cuhk.edu.hk, \{sdongae, danxu\}@cse.ust.hk}
}

\begin{document}

\maketitle

\begin{abstract}
Part-based 3D generation holds great potential for various applications. Previous part generators that represent parts using implicit vector-set tokens often suffer from insufficient geometric details. Another line of work adopts an explicit voxel representation but shares a global voxel grid among all parts; this often causes small parts to occupy too few voxels, leading to degraded quality.
In this paper, we propose \textit{FullPart}, a novel framework that combines both implicit and explicit paradigms.
It first derives the bounding box layout through an implicit box vector-set diffusion process, a task that implicit diffusion handles effectively since box tokens contain little geometric detail.
Then, it generates detailed parts, each within its own fixed full-resolution voxel grid. Instead of sharing a global low-resolution space, each part in our method—even small ones—is generated at full resolution, enabling the synthesis of intricate details.
We further introduce a center-point encoding strategy to address the misalignment issue when exchanging information between parts of different actual sizes, thereby maintaining global coherence. Moreover, to tackle the scarcity of reliable part data, we present \textit{PartVerse-XL}, the largest human-annotated 3D part dataset to date with 40K objects and 320K parts. Extensive experiments demonstrate that FullPart achieves state-of-the-art results in 3D part generation. We will release all code, data, and model to benefit future research in 3D part generation. Project page: \url{https://fullpart3d.github.io}
\end{abstract}

\begin{figure}[hb]
    \centering
    \includegraphics[width=\textwidth]{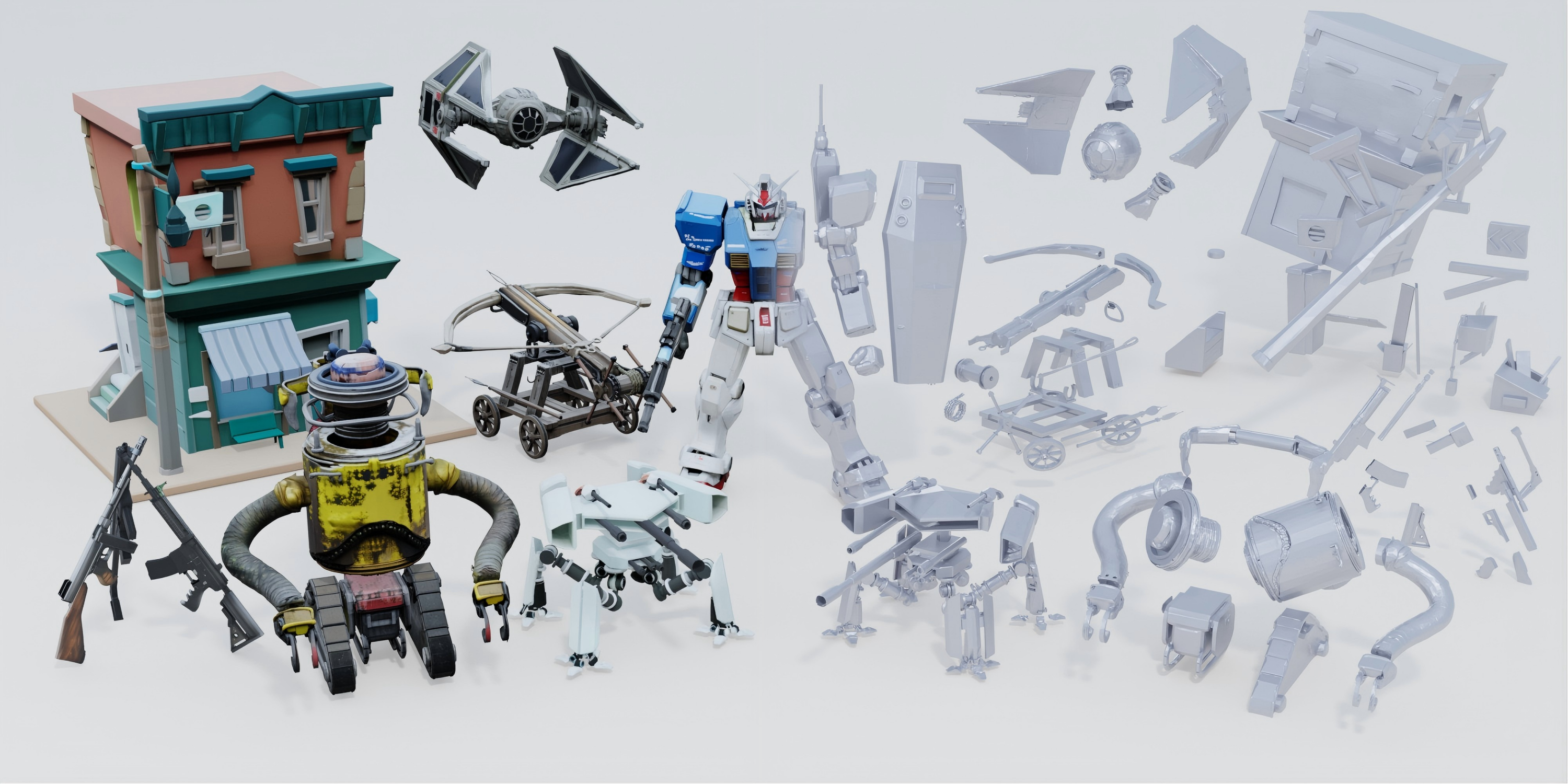}
    \captionof{figure}{FullPart achieves high-quality part-based 3D generation.}
    \label{fig:teaser}
\end{figure}

\section{Introduction}\label{sec:intro}

Part-based 3D generation and manipulation have wide applications in virtual reality, gaming, robotics, and digital content creation. While recent neural 3D generation methods have demonstrated impressive results in 3D object synthesis~\citep{zhang2024clay,xiang2025trellis,li2025triposg,zhao2025hunyuan3d}, the majority of methods did not provide detailed part decomposition, which is essential for downstream tasks such as texture mapping, animation, physical simulation, or fine-grained editing processes. Ideally, a more useful 3D generation framework shall also capture the compositional nature of real-world objects. It shall decompose the generated object into semantically meaningful parts, and users can manipulate each part independently while global coherence can be maintained automatically.

To achieve this, researchers recently proposed part-level synthesis methods, primarily focusing on two paradigms. One paradigm employs implicit latent part-based representations~\citep{lin2025partcrafter,chen2025autopartgen,tang2025efficient}, where each part corresponds to an independent set of latent tokens jointly generated by a shared model. This approach benefits from end-to-end training that simultaneously learns both part geometry and spatial layout, simplifying the training pipeline. However, this implicit representation i) suffers from insufficient part details due to the limited query resolution when decoding part vecsets and ii) cannot precisely model spatial mappings, making it suboptimal for texture generation, multi-modal applications, or precise 3D editing. Another paradigm by~\citet{yang2025omnipart} explicitly defines part layer using bounding boxes, and generates detailed voxel structures within each box. While this approach is good for layout modeling, it is challenging to fine detail generation and maintain the global structural coherence, particularly when handling complex or intricately connected components. Besides, all existing approaches share a critical limitation: both solutions force all parts to share a single global representation space, which limits the resolution allocated to each part and results in poor details of small but complex 3D parts.

In this work, we present two key insights: i) while implicit representations struggle with fine part details, they are well-suited for generating layouts that contain only bounding box information without geometric detail; and ii) explicit representations should allocate an isolated full-resolution space to each part; otherwise, small parts may occupy only a few voxels, resulting in degraded quality.

Based on the these observations, we present FullPart, 
a novel 3D part generation framework that first derives layouts (bounding boxes) from implicit vecset generation and then generates each part at full resolution with explicit representation. In this way, we combine both implicit and explicit paradigms while addressing their respective limitations. Specifically, FullPart follows a three-stage generation process: (1) layout generation by representing bounding boxes with latent vecsets; (2) dividing each box into an isolated $N^3$ grid and generating coarse 3D part structure with full resolution; and (3) refinement of textured meshes based on the coarse structural foundation. 

One remaining challenge of this design is to maintain part coherence when assigning each part to an isolated full-resolution grid. Since each part is generated in its own fixed grid space ($64^3$ in our setup), there is a resolution mismatch problem for tokens on the boundary of two neighboring parts with different sizes: tokens (voxels) from the larger part contain fewer details and tokens from the smaller part contain more finer details. This may result in artifacts in overlapping regions when stitching two parts together. Moreover, since tokens from different parts correspond to voxels of varying spatial sizes within the global coordinate system, directly applying attention mechanisms to exchange information between part tokens would result in scale misalignment.
To address these issues, we propose a specialized center-corner encoding mechanism. For each part, we calculate the positional embeddings of all eight corners and inject them as well as the centers into all tokens in this part. In this way, each token is aware of its actual spatial extent in the voxel grid, and based on which, diffusion will learn to stitch different parts smoothly. Additional analysis also shows this also makes finetuning easier, which we will further discuss in Section~\ref{subsec:coarse}. With this design, the generated 3D parts can be smoothly stitched together, as shown in Figure~\ref{fig:teaser}.
Furthermore, to support high-quality part generation, we introduce PartVerse-XL, the largest and most comprehensively annotated 3D part dataset, consisting 40K objects and 320K parts, with associated part-aware texture descriptions. The reason we built this new dataset is that existing 3D datasets either have no 3D part labeling or very few objects have part labels, or the label quality is not high enough. Some 3D model metadata contains part information from artists' modeling processes, these annotations are often incomplete and lack semantic consistency (e.g., some artists may treat object skin as a separate part). Therefore, we selected 40K objects from Objaverse-XL~\citep{deitke2023objaversexl}, and create a high-quality part annotation, using mesh pre-segmentation followed by human refinement. We will release this dataset to benefit future research in part generation.

Through extensive experimentation, we demonstrate that FullPart achieves superior performance in both part-level fidelity and global structural coherence compared to existing approaches, with particular strength in generating plausible geometries for occluded and small parts.

In summary, our contributions are:
i) We propose FullPart, a novel part-level 3D generation framework that combines implicit layout representation with explicit part structure generation, enabling precise detail control;
ii) We enable each part to be generated in an isolated full resolution while maintaining the global part coherence and preventing violation of the foundation model's established knowledge by proposing a center-corner encoding strategy;
iii) We present PartVerse-XL, a large-scale human-annotated part dataset containing 320K high-quality parts with part-aware textual descriptions, addressing the scarcity of reliable part-level 3D training data;
iv) We demonstrate state-of-the-art performance in part-level 3D generation across multiple metrics, with particular strength in handling complex part interactions and generating plausible occluded geometries.






\section{Related Work}
\subsection{3D Generation}
Early 3D generation focused on category-specific or image-to-3D approaches with limited object diversity~\citep{poole2022dreamfusion,wang2023prolificdreamer,lin2023magic3d,dong2024interactive3d,hong2023lrm,liu2023zero123,shi2023mvdream,ding2024bidiff}. This paradigm shifted with 3D-native diffusion models operating directly in 3D space. CLAY~\citep{zhang2024clay} established a foundational transformer-based architecture for direct 3D generation. Recent advances (2024–2025) significantly improved fidelity and control: TRELLIS~\citep{xiang2025trellis} introduced sparse voxel-based structured latent representations for precise geometry; TripoSG~\citep{li2025triposg} leveraged SDF representations with rectified flow for speed-quality trade-offs; Direct3D~\citep{wu2025direct3ds2} enabled high-resolution text-to-3D generation; and Hunyuan3D~\citep{zhao2025hunyuan3d} integrated multi-modal understanding for text-guided manipulation. However, these frameworks generate monolithic shapes without explicit part decomposition, limiting fine-grained editing.

\subsection{Part Generation}
The need for part-aware 3D generation has motivated several research directions. Early part-aware methods relied on category-specific annotations via auto-encoders (SPAGHETTI by~\citet{hertz2022spaghetti}, Neural Template by \citet{hui2022neural}) or diffusion-based part generation (SALAD by~\citet{koo2023salad}, DiffFacto~\citep{nakayama2023difffacto}), but lacked generalizability. Later works (Part123 by ~\citet{liu2024part123}, PartGen by~\citet{chen2025partgen}) used SAM~\cite{kirillov2023segmentany} for multi-view segmentation yet remained constrained by segmentation quality and limited patch information. However, these methods not only depend critically on segmentation quality but also struggle with the limited information in small segmented patches, constraining reconstruction quality.
Recent contemporaneous works have made significant strides toward general-purpose part generation. CoPart~\citep{dong2025copart} represents a 3D object with multiple contextual part latents and simultaneously generates coherent 3D parts. HoloPart~\citep{yang2025holopart} proposed a holistic part generation framework that jointly optimizes part layout and geometry. PartCrafter~\citep{lin2025partcrafter} introduced an implicit latent representation approach where each part corresponds to an independent set of latent tokens generated by a shared model. AutoPartGen~\citep{chen2025autopartgen} automated part decomposition through a learnable part proposal mechanism. OmniPart~\citep{yang2025omnipart} adopted explicit voxel-based representations with bounding boxes to define part layouts before generating voxel structures within each designated region. BANG~\citep{zhang2025bang} developed an efficient part generation framework based on bounding box attention.

However, all existing part-aware generation methods share a critical limitation: they force parts to share a single global representation space. In implicit representation approaches, this means small parts receive insufficient representation capacity in the shared latent space. In voxel-based methods, this results in small parts occupying only a tiny fraction of a shared N×N×N grid, leading to extremely low effective resolution for those components. This fundamental bottleneck has not been adequately addressed by prior work, which is precisely where our framework makes its key contribution by treating each part as a full-resolution independent object during the generation process.
\begin{figure*}[tbp]
\setlength{\belowcaptionskip}{-0.4cm}
\centering{\includegraphics[width=0.98\linewidth]{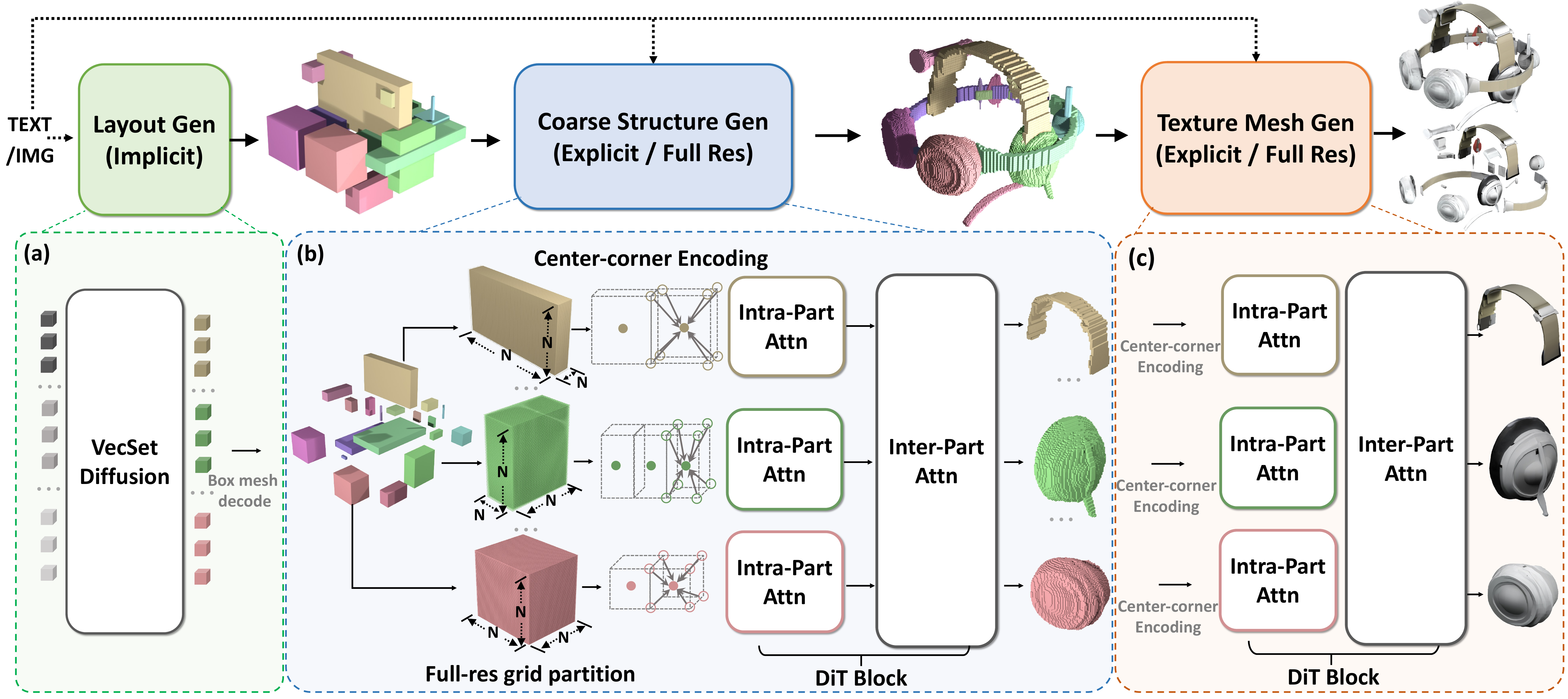}}
\caption{FullPart framework. FullPart comprises three sequential stages: (a) layout generation using implicit vecset diffusion, (b) generating each part at a full-resolution grid with explicit voxel representation, and (c) refining coarse part structures to texture meshes. }
\label{fig:framework}
\end{figure*}

\section{Methodology}\label{sec:method}
\vspace{-2mm}
We formally define our part-aware 3D generation problem as follows: given a conditioning input (typically a single-view RGB image or a text prompt), our goal is to generate a structured 3D object $\mathbf{O} = \{\mathbf{o}_i\}_{i=1}^{K}$ consisting of $K$ semantically meaningful parts, where each part $\mathbf{o}_i$ is represented as a textured mesh with explicit geometric and topological properties. Unlike monolithic 3D generation approaches, our framework explicitly models the compositional nature of objects through a hierarchical generation process that first establishes part layouts (bounding boxes) using implicit vecset diffusion~\citep{li2025triposg} (Figure~\ref {fig:framework} (a)), then generates coarse part structures within boxes by representing each part in an explicit full-resolution voxel grid (Figure~\ref {fig:framework} (b)), and finally refines coarse voxels into detailed meshes with textures (Figure~\ref {fig:framework} (c)). 
In the remainder of this section, we provide a comprehensive description of each component.

\vspace{-0.2cm}
\subsection{Preliminary: 3D Object Generation Frameworks}\label{subsec:preliminary}
\vspace{-2mm}
Our methodology builds upon two predominant paradigms in 3D generation: implicit latent representations for layout generation and explicit voxel-based representations for part structure generation. We briefly review these approaches and establish the foundation for our part-aware extension.

\textbf{Implicit Representations.} Following 3DShape2VecSet~\citep{zhang20233dshape2vecset}, a 3D object can be represented as a vecset—a set of latent tokens $\mathbf{T} = \{\mathbf{t}_j\}_{j=1}^M \in \mathbb{R}^{M \times D}$, where $M$ is the number of tokens and $\mathbf{t}_j \in \mathbb{R}^{D} $ is the $j$-th token with feature dimension $D$. The decoder transforms these tokens into a signed distance field (SDF) representation $\phi: \mathbb{R}^3 \rightarrow \mathbb{R}$, where the zero level set defines the object surface: 
$\mathcal{S} = \{\mathbf{x} \in \mathbb{R}^3 | \phi(\mathbf{x}) = 0\}.$
This implicit representation enables high-fidelity geometry generation but lacks explicit spatial partitioning for part manipulation.

\textbf{Explicit Representations.} TRELLIS~\citep{xiang2025trellis} introduces a structured latent representation for 3D objects through sparse voxels. Given a 3D asset, it encodes geometric information into a set of active featured voxels $\mathbf{C}$:
\begin{equation}
    \mathbf{C} = \{\mathbf{c}_i | \mathbf{c}_i = (\mathbf{f}_i, \mathbf{p}_i)\}_{i=1}^L, \quad \mathbf{f}_i \in \mathbb{R}^D, \quad \mathbf{p}_i \in \{0,1,\ldots,N-1\}^3,
\end{equation}
where $\mathbf{p}_i$ denotes the positional index of an active voxel in the $N \times N \times N$ grid, and $\mathbf{f}_i$ represents the feature vector capturing local geometry and appearance. The active voxels $\mathbf{p}_i$ define the coarse structure, while $\mathbf{f}_i$ encodes fine details.

\textbf{Attention Mechanisms for Part Generation.} Building upon the previous works like CoPart~\citep{dong2025copart}, we introduce two attention mechanisms critical for part-aware generation:

\textit{Intra-Part Attention:} For part $k$, given its token set $\mathbf{T}_k \in \mathbb{R}^{M \times D}$, intra-part attention limits the self-attention computation to the tokens of part $k$:
\begin{equation}
    \mathbf{Q}_k = \mathbf{W}_q \mathbf{T}_k; \  \mathbf{K}_k= \mathbf{W}_k \mathbf{T}_k; \ \mathbf{V}_k=\mathbf{W}_v \mathbf{T}_k
\end{equation}
where $\mathbf{Q}_k, \mathbf{K}_k, \mathbf{V}_k$ are query, key, and value projections of $\mathbf{T}_k$.

\textit{Inter-Part Attention:} Given all part tokens $\mathbf{T} = [\mathbf{T}_1, \ldots, \mathbf{T}_K] \in \mathbb{R}^{KM \times D}$, inter-part attention computes self-attention between all tokens:
\begin{equation}
    \mathbf{Q} = \mathbf{W}_q \mathbf{T}; \  \mathbf{K}= \mathbf{W}_k \mathbf{T}; \ \mathbf{V}=\mathbf{W}_v \mathbf{T}
\end{equation}
where $\mathbf{Q}, \mathbf{K}, \mathbf{V}$ are projections of the concatenated token set $\mathbf{T}$.

These attention mechanisms enable our framework to balance intra-part detail generation with inter-part structural coherence. The flexibility of transformer architecture allows seamless adaptation of full-object generation models to part-level synthesis while preserving pre-trained model priors.

\vspace{-3mm}
\subsection{Layout Generation}\label{subsec:layout}
\vspace{-2mm}
Our layout generation module produces a set of bounding boxes that define the spatial arrangement of object parts. Rather than treating boxes as abstract parameters, we represent each box $\mathbf{b}_k$ as a minimal triangular mesh (a cuboid with 8 vertices and 12 faces), where the collection of these meshes forms a coarse ``blocky" representation of the object, reminiscent of Minecraft-style models.
By representing bounding boxes as meshes, we obtain a semantic representation aligned with the latent space of the vecset diffusion model, thereby leveraging its strong prior for effective layout generation.

Formally, given a conditioning image or text prompt, we generate a set of $K'$ bounding box meshes B = $\{\mathbf{b}_k\}_{k=1}^{K'}$, where $\mathbf{b}_k = (\mathbf{v}_k, \mathbf{f}_k)$ is the $k$-th box mesh with cuboid vertices $\mathbf{v}_k \in \mathbb{R}^{8 \times 3}$ and face indices $\mathbf{f}_k \in \mathbb{N}^{12 \times 3}$.

To encode these box meshes, we utilize the VAE from TripoSG~\citep{li2025triposg}, which maps each box to $M$ latent tokens: $ \mathbf{T}_k = \text{VAE}_{\text{enc}}(\mathbf{b}_k) \in \mathbb{R}^{M \times D}, \quad k = 1,\ldots,K'.$

To make the network distinguish tokens from different boxes, we inject box ID embeddings $\mathbf{e}_{\text{id}}(k) \in \mathbb{R}^D$ to the corresponding token $\mathbf{t}_{k} \in \mathbb{R}^D$ (tokens from box $k$): $\tilde{\mathbf{t}}_{k} = \mathbf{t}_{k} + \mathbf{e}_{\text{id}}(k).$

Additionally, to leverage the powerful priors of the foundation model, we retain a global branch in the original vecset diffusion model, which is tasked with predicting the holistic object structure. This global branch provides essential semantic guidance for the layout generation process. To enable the network to distinguish its tokens from part-specific box tokens, we assign an identifier of 0 to the tokens~$\mathbf{T}_0 \in \mathbb{R}^{M \times D}$ in the global branch.


During training, for objects with $K < K'$ actual boxes, we pad the token sequences with zero vectors for the remaining $K'-K$ boxes. The complete token sequence that represents the box layout for the DiT input is: $\mathbf{T}_{\text{all}} = [\mathbf{T}_0, \mathbf{T}_1, \ldots, \mathbf{T}_{K'}] \in \mathbb{R}^{(K'+1)M \times D},$
where $T_0$ is the tokens from the global branch.
Our DiT employs a hybrid attention strategy: intra-part attention operates within each box's token set ($\mathbf{T}_k$ for $k=0,1,\ldots,K'$), while inter-part attention operates across all box tokens. This enables the model to learn both local-specific characteristics and global structural relationships.

At inference time, we sample the latent tokens $\mathbf{T}_{\text{all}}$ using a diffusion process, then decode them back to box meshes via the VAE decoder: $\hat{\mathbf{B}} = \{\text{VAE}_{\text{dec}}(\mathbf{T}_k)\}_{k=1}^{K}.$
Finally, due to potential deformation and incorrect shapes with the decoded meshes $\hat{\mathbf{B}}$, we recalculate their own bounding boxes $\hat{\mathbf{B}}'$, and retain those with high IoU between corresponding $\hat{\mathbf{B}}$ and $\hat{\mathbf{B}}'$ as layout inputs for subsequent stages.

\vspace{-3mm}
\subsection{3D Coarse Structure Generation}\label{subsec:coarse}
\begin{figure*}[tbp]
\setlength{\abovecaptionskip}{0.1cm}
\setlength{\belowcaptionskip}{-0.6cm}
\centering{\includegraphics[width=0.98\linewidth]{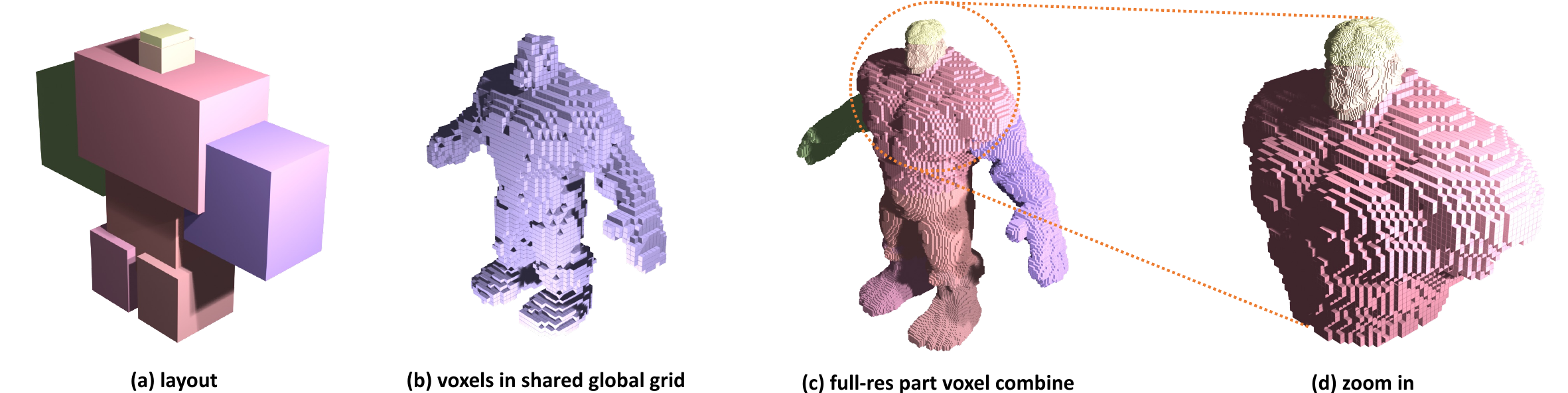}}
\caption{Illustration of our 3D part representation. Our model generates each part at isolated full resolution (c), which contains more fine details than the previous sharing global voxel grid strategy (b). Also, tokens from different parts represent varying spatial extents, e.g., head and body in (d).}
\label{fig:misalign}
\end{figure*}

\vspace{-2mm}
Given the bounding boxes from the layout generation, our coarse structure generation stage creates detailed part geometries within the layout-defined regions. Instead of sharing a global voxel grid (Figure~\ref{fig:misalign} (b)), we generate each part within its own dedicated $N \times N \times N$ voxel space at full resolution, enabling precise detail generation regardless of part size, as shown in Figure~\ref{fig:misalign} (c).

To generate each part at a full resolution, we normalize it to the canonical space $[-1,1]^3$ and define a binary occupancy grid: $\mathbf{V}_k \in \{0,1\}^{N^3}, \text{where } \mathbf{V}_k[x,y,z] = 1 \text{ if occupied},$
and $k$ denotes the $k$-th part.
This normalization ensures that even small parts utilize the full resolution of the voxel grid, overcoming the resolution limitations of shared-grid approaches.

However, when each part is defined in its own full-resolution grid, the tokens from different bounding box regions represent varying spatial extents. This discrepancy becomes problematic when exchanging information between parts. For example, a token from a big part (e.g., upper body in Figure~\ref{fig:misalign} (d)) may represent an absolute voxel size several times larger than a token from a small part (e.g., head in Figure~\ref{fig:misalign} (d)), leading to token misalignment.

To address this, we introduce a \textit{center-corner encoding mechanism} that embeds the absolute spatial context of each voxel. For a voxel at position $\mathbf{u} = (x,y,z)$ in the normalized grid of part $k$, we compute its 8 corresponding corners in the global object space: $\{\mathbf{u}_g^{i} | \mathbf{u}_g^{i} = \mathcal{T}(\mathbf{u}^{i}, \mathbf{b}_k)\}_{i=0}^{7}$, where $\mathcal{T}(,\mathbf{b}_k)$ is the transformation to global coordinate using box $\mathbf{b}_k$ and $\mathbf{u}_g^{i}$ is the $i$-th corner of $\mathbf{u}$ in global coordinate. Following the foundation model that encodes integer positions in the global coordinate, we partition global space into a super-high resolution grid ($2048 \times 2048 \times 2048$) and find the integer coordinates of all eight corners $\{\lfloor{\mathbf{u}_g^{i}}\rfloor\}_{i=0}^{7}$ and the center $\lfloor{\mathbf{u}_g}\rfloor$.

At this stage, we have obtained the center and eight corner coordinates for each voxel of every part. Although different voxels may represent different spatial extents, their corner and center positions are now expressed within a unified super-high-resolution global coordinate system. This allows us to encode these positions using the pre-trained positional embedding layer directly.

One remaining consideration is that our model is based on a  pre-trained 3D generator, whose positional encoding layers were only trained under a low resolution of $64 \times 64 \times 64$. Fortunately, previous research has shown that positional encoding can be effectively extrapolated during fine-tuning~\citep{liu2023scaling}. Therefore, we simply inject the positional embeddings of one center and eight corners for each token: $\mathbf{t}_\mathbf{u}^k = \mathbf{e}_{\text{pos}}(\lfloor{\mathbf{u}_g}\rfloor) + \sum_{i=0}^{7}\mathbf{e}_{\text{pos}}(\lfloor{\mathbf{u}_g^{i}}\rfloor) + \mathbf{e}_{\text{id}}(k),$
where $\mathbf{e}_{\text{pos}}()$ is the positional embedding layer and $\mathbf{e}_{\text{id}}()$ is an additional embedding layer to inject part ID information.
This center-corner embedding is added to each voxel token, providing explicit spatial context that enables the model to understand part relationships despite the normalization.
Furthermore, our strategy requires no modification to the pre-trained model's architecture, thus it can better utilize its foundational priors.

Similar to the layout stage, we include a global branch ($k=0$) that processes the entire object structure. The complete token sequence is: $\mathbf{T}_{\text{all}} = [\mathbf{T}_0, \mathbf{T}_1, \ldots, \mathbf{T}_{K'}].$

Our DiT architecture employs the same hybrid attention mechanism as in the layout stage, allowing it to capture both fine-grained part details and global structural coherence. The diffusion process generates the voxel tokens conditioned on the input image or text prompt, which is incorporated through cross-attention layers following TRELLIS's conditioning strategy.

\vspace{-0.4cm}
\subsection{Refinement}\label{subsec:refinement}
\vspace{-3mm}
The refinement stage enhances the coarse voxel structures with detailed geometry and textures, adapting TRELLIS's second stage~\citep{xiang2025trellis} for part-aware generation. For each part $k$, we obtain feature vectors $\mathbf{F}_k \in \mathbb{R}^{L_k \times D_f}$  for the occupied voxels $\mathcal{P}_k$ via TRELLIS's VAE encoder, where $D_f$ is the feature dimension.

The VAE encoder embeds the tokens with both structure and projected multi-view features (from Dino-v2 by~\citet{oquab2023dinov2}). A key challenge is handling occlusions between parts. To address this, we normalize and render each part independently, obtaining tokens with part-specific image features, as the denoising target during training. However, during inference, our model only requires a single global image or text as a conditional input, making it practical for real-world applications.

The diffusion model generates the feature vectors $\mathbf{F}_k$ on all occupied voxels $\mathcal{P}_k$ at full resolution. These features are then decoded into textured meshes using TRELLIS's decoder: $\hat{\mathbf{o}}_k = \text{Decoder}(\mathbf{F}_k, \mathcal{P}_k).$
The final object is assembled from the individual part meshes: $\hat{\mathbf{O}} = \bigcup_{k=1}^{K} \hat{\mathbf{o}}_k$.

\vspace{-0.3cm}
\subsection{Optimization Loss}\label{subsec:loss}
\vspace{-0.2cm}

All stages of our framework are trained using Conditional Flow Matching (CFM)~\citep{lipman2022flow} objective, i.e., for a given stage with token representations $\mathbf{x}$, we have the training objective:
\begin{equation}
    \mathcal{L}_{\text{cfm}}(\boldsymbol{\theta}) = \mathbb{E}_{\mathbf{x}_{0}, \epsilon, t} \left[ \|\mathbf{v}_{\theta}(\mathbf{x},t) - (\boldsymbol{\epsilon} - \mathbf{x}_{0}) \|^2_2 \right].
\end{equation}
where $\epsilon_{\theta}$ is the diffusion noise, and $\mathbf{v}_{\theta}$ is the predicting vector field.

\vspace{-0.4cm}
\section{PartVerse-XL Dataset}\label{subsec:dataset}
\vspace{-0.4cm}

To support large-scale part-aware 3D generation, we introduce \textit{PartVerse-XL}, an expanded and refined extension of \textit{PartVerse}~\citep{dong2025copart}. It contains 40K high-quality 3D objects from \textit{Objaverse-XL}~\citep{deitke2023objaversexl}, yielding 320K semantically consistent, textured parts across over 200 categories, each paired with a descriptive caption. This scale and diversity significantly surpass prior benchmarks and enable robust training of models requiring fine-grained part semantics and geometry.

We construct \textit{PartVerse-XL} via a two-stage pipeline. First, we apply an automated pre-segmentation algorithm that fuses geometric priors (e.g., mesh connectivity, UV seams) with semantic cues from SAM-2~\citep{kirillov2023segmentany} and Samesh~\citep{tang2024samesh}, deliberately producing over-segmented outputs for easier human correction. Second, expert annotators refine the segments using a Blender-based tool—merging or splitting components to ensure semantic clarity, structural symmetry, and texture preservation—while discarding low-quality or ambiguous assets.

For each part, we generate a textual caption by rendering multi-view images of both the full object and the isolated part. We select the view with maximal part–whole visibility overlap, overlay a bounding box around the part, and feed the composite image to a vision-language model. Captions are generated to accurately describe shape, appearance, material, and part–object relationships (e.g., ``a cylindrical metallic handle attached to the right side of a coffee mug").

\textit{PartVerse-XL} establishes a new standard in scale, semantic fidelity, and multimodal alignment for part-level 3D generation.
More details of our dataset can be found in the supplementary material.


\begin{figure*}[tbp]
\setlength{\abovecaptionskip}{-0.1cm}
\setlength{\belowcaptionskip}{-0.4cm}
\centering{\includegraphics[width=\linewidth]{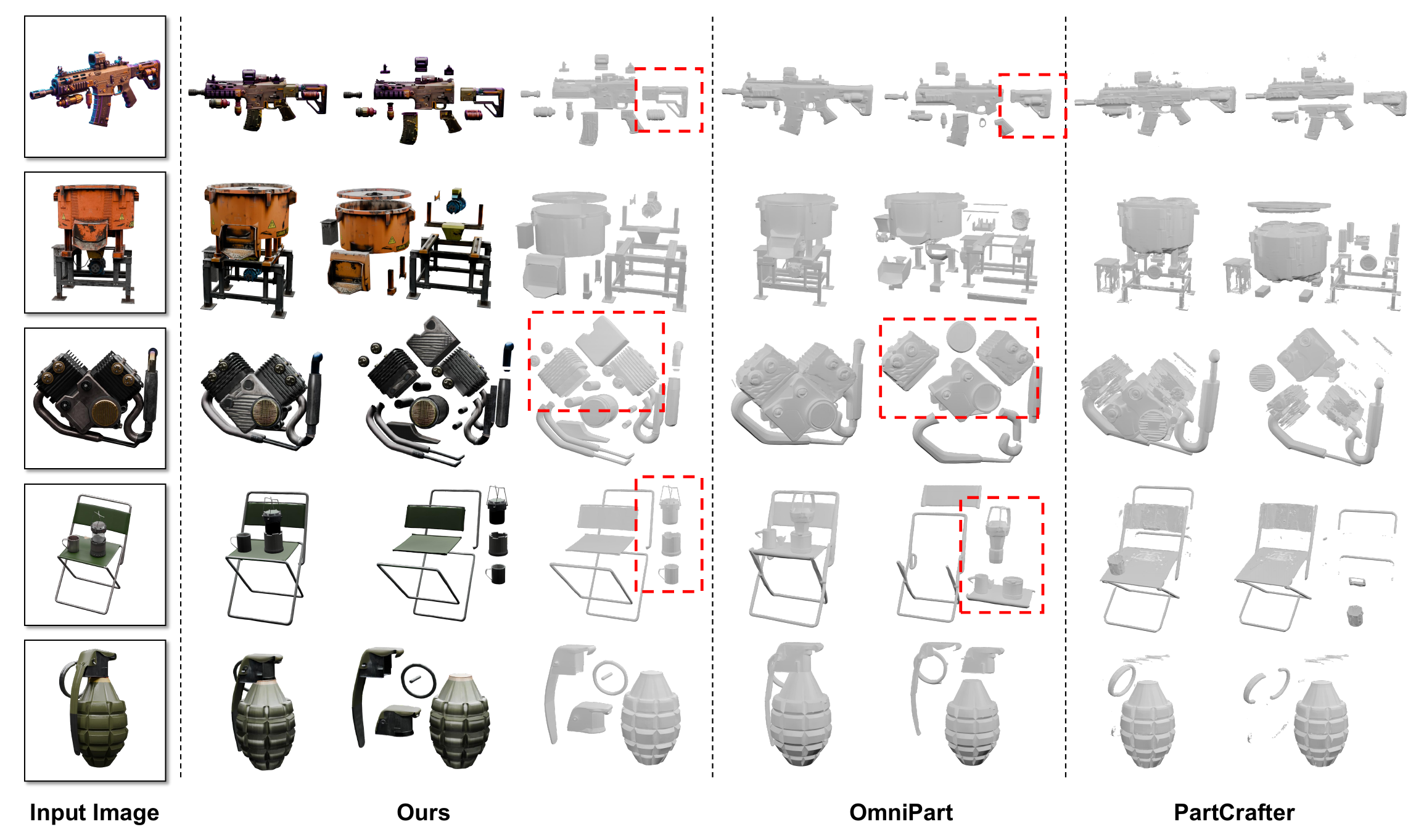}}
\vspace{-1em}
\caption{Comparison with state-of-the-art 3D Part generators. Our method can generate more detailed and reasonably divided parts.}
\label{fig:comp_sota}
\vspace{-.3em}
\end{figure*}

\vspace{-0.4cm}
\section{Experiments}\label{sec:experiments}
\vspace{-0.3cm}

This section presents our experimental results. All experiments of our model utilize the PartVerse-XL training set (40K objects, 320K parts), and we also construct a dedicated test set of 100 objects.

\vspace{-0.3cm}
\subsection{Implementation Details}\label{sec:impl_details}
\vspace{-2mm}
We train FullPart in three sequential stages on 8 NVIDIA A100 GPUs. The layout generator (Stage 1) is trained for 96 hours using AdamW ($\beta_1=0.9$, $\beta_2=0.999$) with batch size 64. Stages 2 (coarse voxel generation) and 3 (mesh refinement) each train for 144 hours with batch size 8, leveraging pre-trained TRELLIS~\cite{xiang2025trellis} weights. The maximum part count is clamped to $K_{\max}=30$, and all parts are generated in isolated full resolution $64^3$ grids. During inference, we apply non-maximum suppression (NMS) with IoU threshold 0.7 to eliminate redundant boxes. Besides, we take 100 manually selected untrained objects as the test set.
More details can be found in the supplementary material.

\vspace{-3mm}
\subsection{Results}\label{sec:exp_results}
\vspace{-2mm}

\subsubsection{Comparison with State-of-the-Art Part Generators}
\vspace{-0.2cm}
We compare FullPart against two leading part-aware generators: PartCrafter~\cite{lin2025partcrafter}, and Omnipart~\cite{yang2025omnipart}. \Cref{fig:comp_sota} demonstrates that FullPart achieves superior geometric fidelity and structural coherence, particularly for intricate assemblies (e.g., articulated robot arms) and occluded regions (e.g., chair undersides). While PartCrafter produces fragmented parts due to implicit token entanglement, and Omnipart suffers from voxelization artifacts in small components (e.g., thin chair legs), FullPart preserves fine details through dedicated per-part full-resolution grids.

\vspace{-3mm}
\subsubsection{Comparison with Full-Object Methods}
\begin{figure*}[tbp]
\setlength{\abovecaptionskip}{0.1cm}
\setlength{\belowcaptionskip}{-0.6cm}
\centering{\includegraphics[width=\linewidth]{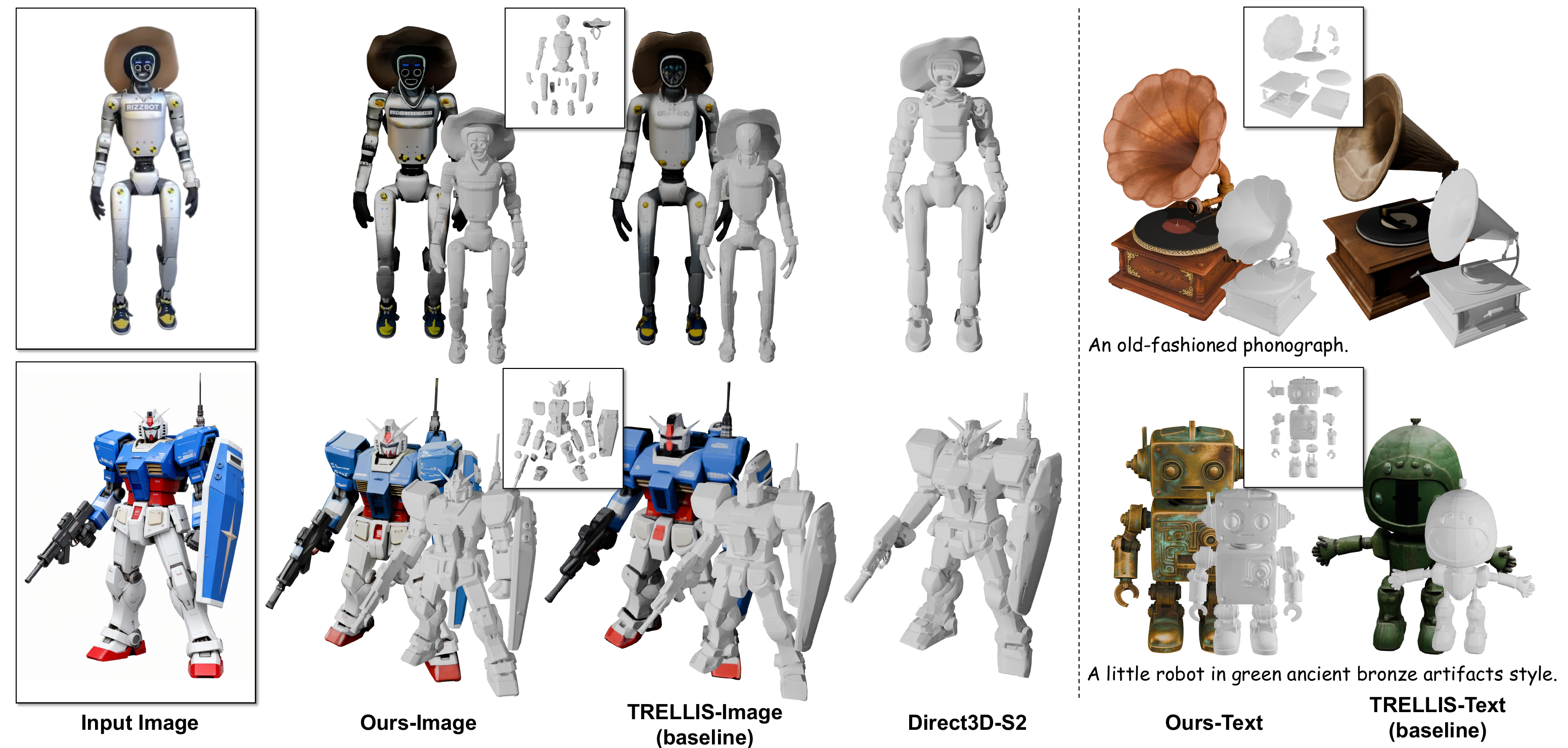}}
\caption{Comparison with the state-of-the-art 3D generators.}
\label{fig:comp_full}
\end{figure*}

\vspace{-0.2cm}
We compare FullPart against its foundational model TRELLIS~\cite{xiang2025trellis} and the state-of-the-art monolithic 3D generator Direct3D-S2~\cite{wu2025direct3ds2}. As these full-object approaches lack part decomposition capability, they inherently suffer from global grid sparsity, leading to significant detail loss in fine-grained regions (e.g., robotic head features in \Cref{fig:comp_full}). In contrast, FullPart's part-aware architecture preserves high-fidelity details through localized high-resolution generation.

\vspace{-0.3cm}
\subsubsection{Quantitative Evaluation}
\vspace{-2mm}
\Cref{tab:quantitative} reports metrics on the 100-object test set. We evaluate: (i) Global fidelity with a threshold of 0.1 (F-Score), (ii) Global mesh chamfer distance (CD), (iii) Part mesh chamfer distance when use same layout boxes (Part-CD), and (iv) 3D Semantic alignment (ULIP Score~\cite{xue2023ulip}). FullPart outperforms all baselines in part-level and full-level metrics, proving its strength in part-level detail and global coherence. 
Note that Part-CD is not applicable to the first three methods due to their lack of bounding box-conditioned part generation capability.

\begin{table}[t]
\centering
\caption{Quantitative comparison on PartVerse-XL test set (``-" denotes not applicable).}
\vspace{-8pt}
\label{tab:quantitative}
\begin{tabular}{l|c|c|c|c}
\hline
\textbf{Method} & \textbf{F-Score} $\uparrow$ & \textbf{CD} $\downarrow$ & \textbf{Part-CD} $\downarrow$ & \textbf{ULIP-Score} $\uparrow$ \\
\hline
TRELLIS~\cite{xiang2025trellis} & 0.71 & 0.16 & - & 0.21 \\
HoloPart~\cite{lin2025partcrafter} & 0.68 & 0.21 & - & 0.15 \\
PartCrafter~\cite{chen2025autopartgen} & 0.63 & 0.42 & - & 0.13 \\
Omnipart~\cite{yang2025omnipart} & 0.77 & 0.15 & 0.42 & 0.22 \\
\hline
\textbf{FullPart (Ours)} & \textbf{0.81} & \textbf{0.11} & \textbf{0.36} & \textbf{0.24} \\
\hline
\end{tabular}
\vspace{-3mm}
\end{table}

\begin{figure*}[tbp]
\setlength{\belowcaptionskip}{-0.4cm}
\centering{\includegraphics[width=\linewidth]{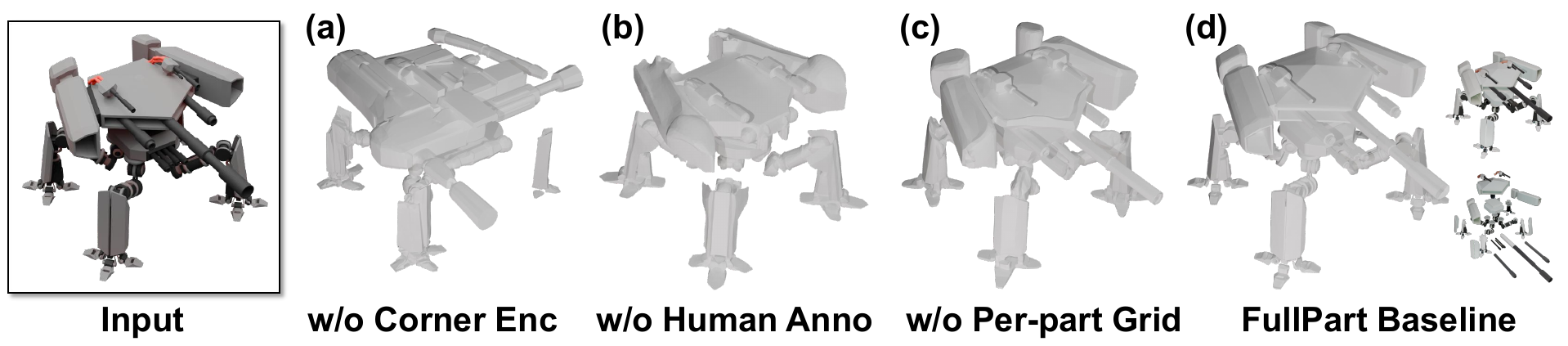}}
\vspace{-1em}
\vspace{-0.3cm}
\caption{Comparison with different model settings under identical training budgets: (a) no corner encoding, (b) using metadata-derived structural information without manual annotations, (c) all layout boxes constrained to a single voxel space, causing each box may only occupy a small number of voxels, and (d) normal setting with all things.}
\label{fig:ab1}
\vspace{-.3em}
\end{figure*}

\vspace{-0.3cm}
\subsection{Ablation Studies}\label{sec:exp_ablation}
\vspace{-2mm}
\textbf{\noindent{Center-corner Encoding Ablation.}}
To validate our center-corner encoding strategy, we ablate the generator by replacing center + corner coordinates with only center coordinates. \Cref{fig:ab1} (a) shows that this variant fails to model part interactions (e.g., misaligned chair legs). This confirms that explicit location and scale information are critical for capturing spatial relationships between parts.

\textbf{\noindent{Impact of Human-post Annotations.}}
We train FullPart on two data variants: (i) raw metadata (artist-provided part labels), and (ii) PartVerse-XL with human-refined annotations. As \Cref{fig:ab1} (b) (d) illustrates, metadata-only training produces semantically incorrect parts due to the noise in part labels, while training on human-annotated data yields coherent, functionally meaningful parts. 

\textbf{\noindent{Per-Part Full-Resolution Grid Ablation.}}
We compare our dedicated $N^3$ ($N=64$ in our setting) per-part grids against a global grid baseline where all parts share the global $N^3$ space. 
\Cref{fig:ab1} (c) reveals severe detail degradation in small parts under the global grid, as they occupy only a small number of voxels. Our method maintains consistent resolution across all parts, achieving uniform detail generation regardless of relative size. 

\vspace{-2mm}
\subsection{Application}
\vspace{-2mm}
\begin{wrapfigure}{r}{0.5\textwidth}
  \setlength{\abovecaptionskip}{-0.0mm}  
  \setlength{\belowcaptionskip}{-4mm}
  \vspace{-3mm}
  \begin{center}
    \includegraphics[width=0.48\textwidth]{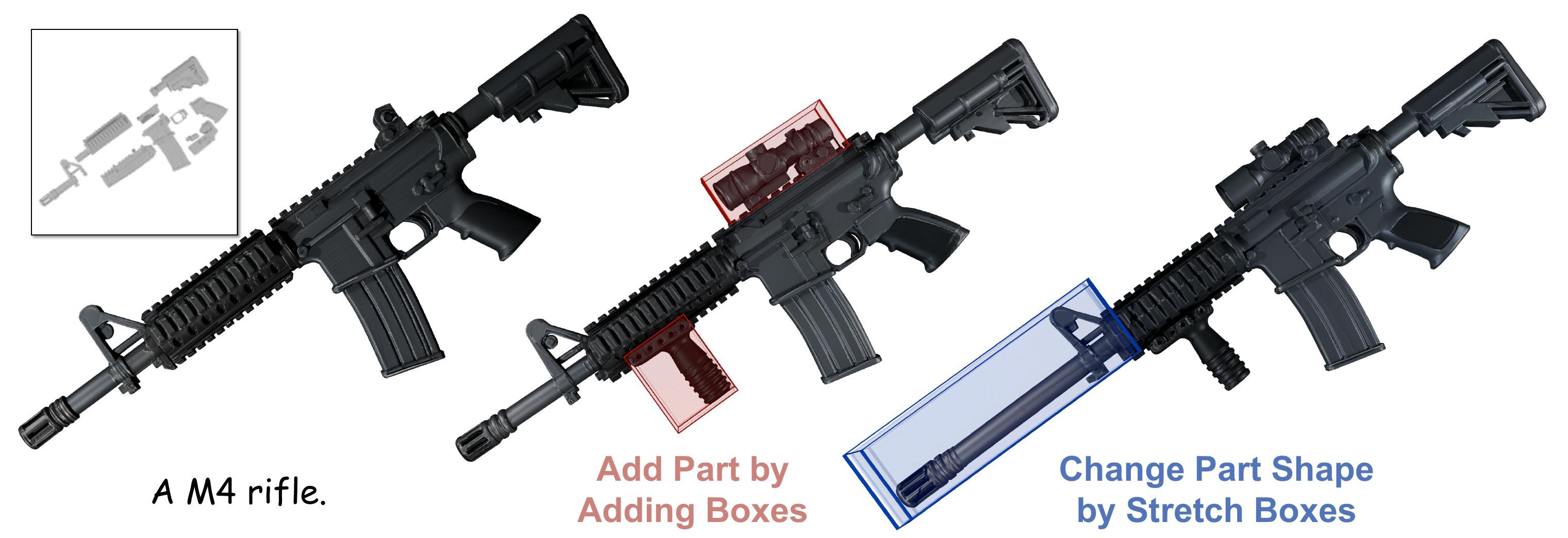}
  \end{center}
  \caption{Editing applications.}
  \label{fig:editing}
\end{wrapfigure}
The FullPart framework exhibits strong flexibility, enabling a range of controllable 3D editing applications through manipulation of the layout boxes. Specifically, users can intuitively edit part-level geometry by adding, deleting, or modifying the shape and position of bounding boxes—each operation directly influences the corresponding generated part while preserving the integrity of the rest of the object. As illustrated in~\Cref{fig:editing}, we demonstrate an example where two accessories are added to a rifle and the barrel is elongated. After the initial generation, we modify the layout boxes by inserting new boxes and stretching existing ones, then re-run inference. For unchanged parts, we bypass the diffusion sampling process by directly injecting their clean latent tokens (along with the appropriate noise level corresponding to the current inference timestep) into the DiT as fixed inputs. This strategy ensures consistent regeneration of unedited components while allowing efficient, localized updates—highlighting FullPart’s suitability for interactive, part-aware 3D content creation.

\vspace{-3mm}
\section{Conclusion}
\vspace{-2mm}
We introduced FullPart, a part-aware 3D generation framework that integrates implicit and explicit representations. By treating each part as a full-resolution object, our method overcomes the resolution bottleneck of shared voxel grids.  We also introduce PartVerse-XL, a large-scale annotated part dataset to advance future research in 3D part generation.

\bibliography{iclr2026_conference}

\begin{thebibliography}{36}
\providecommand{\natexlab}[1]{#1}
\providecommand{\url}[1]{\texttt{#1}}
\expandafter\ifx\csname urlstyle\endcsname\relax
  \providecommand{\doi}[1]{doi: #1}\else
  \providecommand{\doi}{doi: \begingroup \urlstyle{rm}\Url}\fi

\bibitem[Chen et~al.(2025{\natexlab{a}})Chen, Shapovalov, Laina, Monnier, Wang, Novotny, and Vedaldi]{chen2025partgen}
Minghao Chen, Roman Shapovalov, Iro Laina, Tom Monnier, Jianyuan Wang, David Novotny, and Andrea Vedaldi.
\newblock Partgen: Part-level 3d generation and reconstruction with multi-view diffusion models.
\newblock In \emph{CVPR}, 2025{\natexlab{a}}.

\bibitem[Chen et~al.(2025{\natexlab{b}})Chen, Wang, Shapovalov, Monnier, Jung, Wang, Ranjan, Laina, and Vedaldi]{chen2025autopartgen}
Minghao Chen, Jianyuan Wang, Roman Shapovalov, Tom Monnier, Hyunyoung Jung, Dilin Wang, Rakesh Ranjan, Iro Laina, and Andrea Vedaldi.
\newblock Autopartgen: Autogressive 3d part generation and discovery.
\newblock \emph{arXiv preprint arXiv:2507.13346}, 2025{\natexlab{b}}.

\bibitem[Deitke et~al.(2023)Deitke, Liu, Wallingford, Ngo, Michel, Kusupati, Fan, Laforte, Voleti, Gadre, et~al.]{deitke2023objaversexl}
Matt Deitke, Ruoshi Liu, Matthew Wallingford, Huong Ngo, Oscar Michel, Aditya Kusupati, Alan Fan, Christian Laforte, Vikram Voleti, Samir~Yitzhak Gadre, et~al.
\newblock Objaverse-xl: A universe of 10m+ 3d objects.
\newblock In \emph{NeuIPS}, 2023.

\bibitem[Ding et~al.(2024)Ding, Dong, Huang, Wang, Zhang, Gong, Xu, and Xue]{ding2024bidiff}
Lihe Ding, Shaocong Dong, Zhanpeng Huang, Zibin Wang, Yiyuan Zhang, Kaixiong Gong, Dan Xu, and Tianfan Xue.
\newblock Text-to-3d generation with bidirectional diffusion using both 2d and 3d priors.
\newblock In \emph{Proceedings of the IEEE/CVF Conference on Computer Vision and Pattern Recognition}, pp.\  5115--5124, 2024.

\bibitem[Dong et~al.(2024)Dong, Ding, Huang, Wang, Xue, and Xu]{dong2024interactive3d}
Shaocong Dong, Lihe Ding, Zhanpeng Huang, Zibin Wang, Tianfan Xue, and Dan Xu.
\newblock Interactive3d: Create what you want by interactive 3d generation.
\newblock In \emph{CVPR}, 2024.

\bibitem[Dong et~al.(2025)Dong, Ding, Chen, Li, Wang, Wang, Wang, Kim, Gao, Huang, et~al.]{dong2025copart}
Shaocong Dong, Lihe Ding, Xiao Chen, Yaokun Li, Yuxin Wang, Yucheng Wang, Qi~Wang, Jaehyeok Kim, Chenjian Gao, Zhanpeng Huang, et~al.
\newblock From one to more: Contextual part latents for 3d generation.
\newblock In \emph{ICCV}, 2025.

\bibitem[Hertz et~al.(2022)Hertz, Perel, Giryes, Sorkine-Hornung, and Cohen-Or]{hertz2022spaghetti}
Amir Hertz, Or~Perel, Raja Giryes, Olga Sorkine-Hornung, and Daniel Cohen-Or.
\newblock Spaghetti: Editing implicit shapes through part aware generation.
\newblock \emph{ACM Transactions on Graphics (TOG)}, 41\penalty0 (4):\penalty0 1--20, 2022.

\bibitem[Hong et~al.(2023)Hong, Zhang, Gu, Bi, Zhou, Liu, Liu, Sunkavalli, Bui, and Tan]{hong2023lrm}
Yicong Hong, Kai Zhang, Jiuxiang Gu, Sai Bi, Yang Zhou, Difan Liu, Feng Liu, Kalyan Sunkavalli, Trung Bui, and Hao Tan.
\newblock Lrm: Large reconstruction model for single image to 3d.
\newblock \emph{arXiv preprint arXiv:2311.04400}, 2023.

\bibitem[Hui et~al.(2022)Hui, Li, Hu, and Fu]{hui2022neural}
Ka-Hei Hui, Ruihui Li, Jingyu Hu, and Chi-Wing Fu.
\newblock Neural template: Topology-aware reconstruction and disentangled generation of 3d meshes.
\newblock In \emph{CVPR}, 2022.

\bibitem[Kirillov et~al.(2023)Kirillov, Mintun, Ravi, Mao, Rolland, Gustafson, Xiao, Whitehead, Berg, Lo, et~al.]{kirillov2023segmentany}
Alexander Kirillov, Eric Mintun, Nikhila Ravi, Hanzi Mao, Chloe Rolland, Laura Gustafson, Tete Xiao, Spencer Whitehead, Alexander~C Berg, Wan-Yen Lo, et~al.
\newblock Segment anything.
\newblock In \emph{Proceedings of the IEEE/CVF international conference on computer vision}, pp.\  4015--4026, 2023.

\bibitem[Koo et~al.(2023)Koo, Yoo, Nguyen, and Sung]{koo2023salad}
Juil Koo, Seungwoo Yoo, Minh~Hieu Nguyen, and Minhyuk Sung.
\newblock Salad: Part-level latent diffusion for 3d shape generation and manipulation.
\newblock In \emph{ICCV}, 2023.

\bibitem[Li et~al.(2025)Li, Zou, Liu, Wang, Liang, Yu, Liu, Guo, Liang, Ouyang, et~al.]{li2025triposg}
Yangguang Li, Zi-Xin Zou, Zexiang Liu, Dehu Wang, Yuan Liang, Zhipeng Yu, Xingchao Liu, Yuan-Chen Guo, Ding Liang, Wanli Ouyang, et~al.
\newblock Triposg: High-fidelity 3d shape synthesis using large-scale rectified flow models.
\newblock \emph{arXiv preprint arXiv:2502.06608}, 2025.

\bibitem[Lin et~al.(2023)Lin, Gao, Tang, Takikawa, Zeng, Huang, Kreis, Fidler, Liu, and Lin]{lin2023magic3d}
Chen-Hsuan Lin, Jun Gao, Luming Tang, Towaki Takikawa, Xiaohui Zeng, Xun Huang, Karsten Kreis, Sanja Fidler, Ming-Yu Liu, and Tsung-Yi Lin.
\newblock Magic3d: High-resolution text-to-3d content creation.
\newblock In \emph{CVPR}, 2023.

\bibitem[Lin et~al.(2025)Lin, Lin, Pan, Yan, Feng, Mu, and Fragkiadaki]{lin2025partcrafter}
Yuchen Lin, Chenguo Lin, Panwang Pan, Honglei Yan, Yiqiang Feng, Yadong Mu, and Katerina Fragkiadaki.
\newblock Partcrafter: Structured 3d mesh generation via compositional latent diffusion transformers.
\newblock \emph{arXiv preprint arXiv:2506.05573}, 2025.

\bibitem[Lipman et~al.(2023)Lipman, Chen, Ben-Hamu, Nickel, and Le]{lipman2022flow}
Yaron Lipman, Ricky~TQ Chen, Heli Ben-Hamu, Maximilian Nickel, and Matt Le.
\newblock Flow matching for generative modeling.
\newblock In \emph{ICLR}, 2023.

\bibitem[Liu et~al.(2024)Liu, Lin, Liu, Long, Dou, Guo, Luo, and Wang]{liu2024part123}
Anran Liu, Cheng Lin, Yuan Liu, Xiaoxiao Long, Zhiyang Dou, Hao-Xiang Guo, Ping Luo, and Wenping Wang.
\newblock Part123: part-aware 3d reconstruction from a single-view image.
\newblock In \emph{ACM SIGGRAPH 2024 Conference Papers}, 2024.

\bibitem[Liu et~al.(2023{\natexlab{a}})Liu, Wu, Van~Hoorick, Tokmakov, Zakharov, and Vondrick]{liu2023zero123}
Ruoshi Liu, Rundi Wu, Basile Van~Hoorick, Pavel Tokmakov, Sergey Zakharov, and Carl Vondrick.
\newblock Zero-1-to-3: Zero-shot one image to 3d object.
\newblock In \emph{Proceedings of the IEEE/CVF international conference on computer vision}, pp.\  9298--9309, 2023{\natexlab{a}}.

\bibitem[Liu et~al.(2023{\natexlab{b}})Liu, Yan, Zhang, An, Qiu, and Lin]{liu2023scaling}
Xiaoran Liu, Hang Yan, Shuo Zhang, Chenxin An, Xipeng Qiu, and Dahua Lin.
\newblock Scaling laws of rope-based extrapolation.
\newblock \emph{arXiv preprint arXiv:2310.05209}, 2023{\natexlab{b}}.

\bibitem[Mo et~al.(2018)Mo, Zhu, Chang, Yi, Tripathi, Guibas, and Su]{mo2018partnetlargescalebenchmarkfinegrained}
Kaichun Mo, Shilin Zhu, Angel~X. Chang, Li~Yi, Subarna Tripathi, Leonidas~J. Guibas, and Hao Su.
\newblock Partnet: A large-scale benchmark for fine-grained and hierarchical part-level 3d object understanding, 2018.
\newblock URL \url{https://arxiv.org/abs/1812.02713}.

\bibitem[Nakayama et~al.(2023)Nakayama, Uy, Huang, Hu, Li, and Guibas]{nakayama2023difffacto}
George~Kiyohiro Nakayama, Mikaela~Angelina Uy, Jiahui Huang, Shi-Min Hu, Ke~Li, and Leonidas Guibas.
\newblock Difffacto: Controllable part-based 3d point cloud generation with cross diffusion.
\newblock In \emph{CVPR}, 2023.

\bibitem[Oquab et~al.(2023)Oquab, Darcet, Moutakanni, Vo, Szafraniec, Khalidov, Fernandez, Haziza, Massa, El-Nouby, et~al.]{oquab2023dinov2}
Maxime Oquab, Timoth{\'e}e Darcet, Th{\'e}o Moutakanni, Huy Vo, Marc Szafraniec, Vasil Khalidov, Pierre Fernandez, Daniel Haziza, Francisco Massa, Alaaeldin El-Nouby, et~al.
\newblock Dinov2: Learning robust visual features without supervision.
\newblock \emph{arXiv preprint arXiv:2304.07193}, 2023.

\bibitem[Poole et~al.(2022)Poole, Jain, Barron, and Mildenhall]{poole2022dreamfusion}
Ben Poole, Ajay Jain, Jonathan~T Barron, and Ben Mildenhall.
\newblock Dreamfusion: Text-to-3d using 2d diffusion.
\newblock \emph{arXiv preprint arXiv:2209.14988}, 2022.

\bibitem[Shi et~al.(2023)Shi, Wang, Ye, Long, Li, and Yang]{shi2023mvdream}
Yichun Shi, Peng Wang, Jianglong Ye, Mai Long, Kejie Li, and Xiao Yang.
\newblock Mvdream: Multi-view diffusion for 3d generation.
\newblock \emph{arXiv preprint arXiv:2308.16512}, 2023.

\bibitem[Tang et~al.(2024)Tang, Zhao, Ford, Benhaim, and Zhang]{tang2024samesh}
George Tang, William Zhao, Logan Ford, David Benhaim, and Paul Zhang.
\newblock Segment any mesh: Zero-shot mesh part segmentation via lifting segment anything 2 to 3d.
\newblock \emph{arXiv e-prints}, pp.\  arXiv--2408, 2024.

\bibitem[Tang et~al.(2025)Tang, Lu, Li, Hao, Li, Wei, Song, Zeng, Liu, and Lin]{tang2025efficient}
Jiaxiang Tang, Ruijie Lu, Zhaoshuo Li, Zekun Hao, Xuan Li, Fangyin Wei, Shuran Song, Gang Zeng, Ming-Yu Liu, and Tsung-Yi Lin.
\newblock Efficient part-level 3d object generation via dual volume packing.
\newblock \emph{arXiv preprint arXiv:2506.09980}, 2025.

\bibitem[Wang et~al.(2024)Wang, Bai, Tan, Wang, Fan, Bai, Chen, Liu, Wang, Ge, et~al.]{wang2024qwen2}
Peng Wang, Shuai Bai, Sinan Tan, Shijie Wang, Zhihao Fan, Jinze Bai, Keqin Chen, Xuejing Liu, Jialin Wang, Wenbin Ge, et~al.
\newblock Qwen2-vl: Enhancing vision-language model's perception of the world at any resolution.
\newblock \emph{arXiv preprint arXiv:2409.12191}, 2024.

\bibitem[Wang et~al.(2023)Wang, Lu, Wang, Bao, Li, Su, and Zhu]{wang2023prolificdreamer}
Zhengyi Wang, Cheng Lu, Yikai Wang, Fan Bao, Chongxuan Li, Hang Su, and Jun Zhu.
\newblock Prolificdreamer: High-fidelity and diverse text-to-3d generation with variational score distillation.
\newblock \emph{NeuIPS}, 2023.

\bibitem[Wu et~al.(2025)Wu, Lin, Zhang, Zeng, Yang, Bao, Qian, Zhu, Cao, Torr, et~al.]{wu2025direct3ds2}
Shuang Wu, Youtian Lin, Feihu Zhang, Yifei Zeng, Yikang Yang, Yajie Bao, Jiachen Qian, Siyu Zhu, Xun Cao, Philip Torr, et~al.
\newblock Direct3d-s2: Gigascale 3d generation made easy with spatial sparse attention.
\newblock \emph{arXiv preprint arXiv:2505.17412}, 2025.

\bibitem[Xiang et~al.(2025)Xiang, Lv, Xu, Deng, Wang, Zhang, Chen, Tong, and Yang]{xiang2025trellis}
Jianfeng Xiang, Zelong Lv, Sicheng Xu, Yu~Deng, Ruicheng Wang, Bowen Zhang, Dong Chen, Xin Tong, and Jiaolong Yang.
\newblock Structured 3d latents for scalable and versatile 3d generation.
\newblock In \emph{CVPR}, 2025.

\bibitem[Xue et~al.(2023)Xue, Gao, Xing, Mart{\'\i}n-Mart{\'\i}n, Wu, Xiong, Xu, Niebles, and Savarese]{xue2023ulip}
Le~Xue, Mingfei Gao, Chen Xing, Roberto Mart{\'\i}n-Mart{\'\i}n, Jiajun Wu, Caiming Xiong, Ran Xu, Juan~Carlos Niebles, and Silvio Savarese.
\newblock Ulip: Learning a unified representation of language, images, and point clouds for 3d understanding.
\newblock In \emph{CVPR}, 2023.

\bibitem[Yang et~al.(2025{\natexlab{a}})Yang, Guo, Huang, Zou, Yu, Li, Cao, and Liu]{yang2025holopart}
Yunhan Yang, Yuan-Chen Guo, Yukun Huang, Zi-Xin Zou, Zhipeng Yu, Yangguang Li, Yan-Pei Cao, and Xihui Liu.
\newblock Holopart: Generative 3d part amodal segmentation.
\newblock \emph{arXiv preprint arXiv:2504.07943}, 2025{\natexlab{a}}.

\bibitem[Yang et~al.(2025{\natexlab{b}})Yang, Zhou, Guo, Zou, Huang, Liu, Xu, Liang, Cao, and Liu]{yang2025omnipart}
Yunhan Yang, Yufan Zhou, Yuan-Chen Guo, Zi-Xin Zou, Yukun Huang, Ying-Tian Liu, Hao Xu, Ding Liang, Yan-Pei Cao, and Xihui Liu.
\newblock Omnipart: Part-aware 3d generation with semantic decoupling and structural cohesion.
\newblock \emph{arXiv preprint arXiv:2507.06165}, 2025{\natexlab{b}}.

\bibitem[Zhang et~al.(2023)Zhang, Tang, Niessner, and Wonka]{zhang20233dshape2vecset}
Biao Zhang, Jiapeng Tang, Matthias Niessner, and Peter Wonka.
\newblock 3dshape2vecset: A 3d shape representation for neural fields and generative diffusion models.
\newblock \emph{ACM Transactions On Graphics (TOG)}, 42\penalty0 (4):\penalty0 1--16, 2023.

\bibitem[Zhang et~al.(2024)Zhang, Wang, Zhang, Qiu, Pang, Jiang, Yang, Xu, and Yu]{zhang2024clay}
Longwen Zhang, Ziyu Wang, Qixuan Zhang, Qiwei Qiu, Anqi Pang, Haoran Jiang, Wei Yang, Lan Xu, and Jingyi Yu.
\newblock Clay: A controllable large-scale generative model for creating high-quality 3d assets.
\newblock \emph{ACM Transactions on Graphics (TOG)}, 43\penalty0 (4):\penalty0 1--20, 2024.

\bibitem[Zhang et~al.(2025)Zhang, Zhang, Jiang, Bai, Yang, Xu, and Yu]{zhang2025bang}
Longwen Zhang, Qixuan Zhang, Haoran Jiang, Yinuo Bai, Wei Yang, Lan Xu, and Jingyi Yu.
\newblock Bang: Dividing 3d assets via generative exploded dynamics.
\newblock \emph{ACM Transactions on Graphics (TOG)}, 44\penalty0 (4):\penalty0 1--21, 2025.

\bibitem[Zhao et~al.(2025)Zhao, Lai, Lin, Zhao, Liu, Yang, Feng, Yang, Zhang, Yang, et~al.]{zhao2025hunyuan3d}
Zibo Zhao, Zeqiang Lai, Qingxiang Lin, Yunfei Zhao, Haolin Liu, Shuhui Yang, Yifei Feng, Mingxin Yang, Sheng Zhang, Xianghui Yang, et~al.
\newblock Hunyuan3d 2.0: Scaling diffusion models for high resolution textured 3d assets generation.
\newblock \emph{arXiv preprint arXiv:2501.12202}, 2025.

\end{thebibliography}
\bibliographystyle{iclr2026_conference}

\appendix
\clearpage
\section*{Appendix}
\appendix
\section{Implementation Details}
\subsection{Network Architecture}
We fine-tune all three-stage models starting from pre-trained holistic 3D generators. Specifically, the layout generation model is initialized from TripoSG~\cite{li2025triposg}, where we replace the original tokens with our proposed box tokens and inject box ID embeddings to adapt the model for layout generation. Similarly, the coarse structure generation model and the mesh refinement model are initialized using Stage 1 and Stage 2 of TRELLIS, respectively.

In all three stages, we convert half of the DiT blocks to inter-part attention blocks, while the remaining blocks retain intra-part attention. To preserve global semantic context and stabilize fine-tuning, we maintain a holistic generation branch (assigned part ID=0), which helps prevent significant deviation from the pre-trained weights. Furthermore, conditional inputs—such as image or text—are incorporated following the original pre-trained architecture via additional cross-attention blocks. Notably, in each block, all part tokens attend to all condition embeddings.

\subsection{Training and Inference Details}
Owing to GPU memory constraints, we set the maximum number of parts per object to 30 during training.
 Ror each object sample, we sort its parts by the bounding box sizes and choose the top 30 largest parts during training. 
 Despite this limitation, our framework supports the generation of objects with more than 30 parts through a sequential sampling strategy during inference. The process is as follows: first, we sample an initial set of 30 parts alongside a global part that captures the overall shape. Then, in a subsequent sampling round, we replace the global tokens with the noisy version of the previous denoised global token at each timestep, maintaining the global structure unchanged, and sample new 30 parts. 
 We follow TRELLIS~\cite{xiang2025trellis} to render 24 conditional images for each object and randomly choose one during training.
 To improve the quality of conditional generation, we adopt common practices from prior work: condition tokens are randomly dropped with a probability of 0.1 during training, and classifier-free guidance with a scale of 3.5 is applied at inference. We also apply a bounding box augmentation strategy in the training of the coarse structure generation model to make the model robust to imperfect bounding box input.

\section{PartVerse-XL Dataset}
\begin{figure*}[tbp]
\setlength{\belowcaptionskip}{-0.4cm}
\centering{\includegraphics[width=1.1\linewidth]{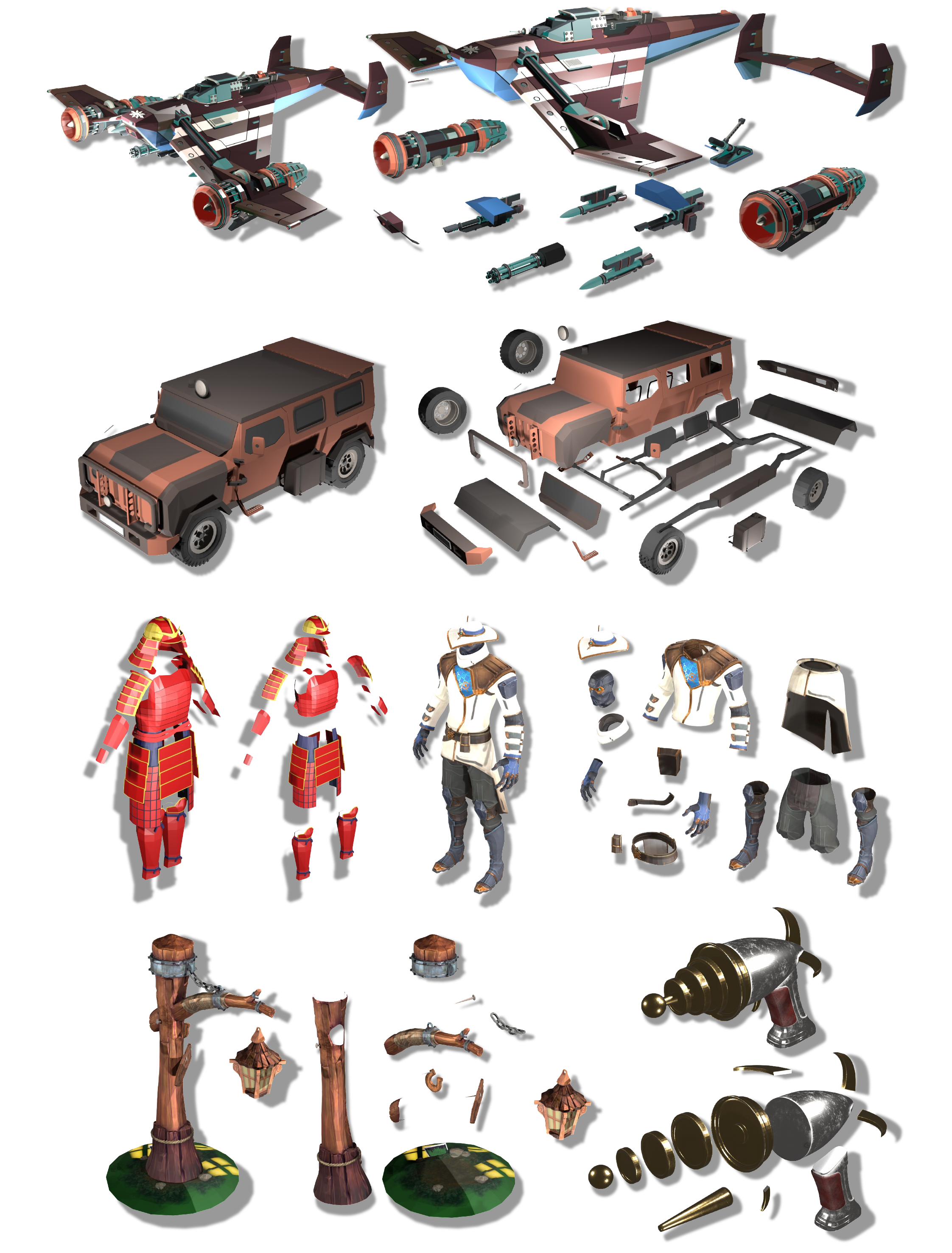}}
\caption{Data examples from PartVerse-XL.}
\label{fig:data_example}
\end{figure*}

To support large-scale, high-fidelity part-aware 3D generation, we present \textit{PartVerse-XL}, a significantly expanded and refined version of the earlier \textit{PartVerse}~\cite{dong2025copart} dataset. Figure.~\ref{fig:data_example} shows some examples in our dataset. \textit{PartVerse-XL} comprises \textbf{40K high-quality 3D objects} sourced from \textit{Objaverse-XL}~\cite{deitke2023objaversexl}, yielding a total of \textbf{320K semantically consistent and texture-preserving parts}, each accompanied by a detailed part-level textual description. This scale and diversity—spanning over 200 object categories—substantially surpasses existing part-level benchmarks such as PartNet~\cite{mo2018partnetlargescalebenchmarkfinegrained} and enables robust training of generative models like FullPart that demand both geometric precision and semantic grounding at the part level.

The construction of \textit{PartVerse-XL} follows a two-stage pipeline that combines automated pre-segmentation with rigorous human refinement, ensuring both scalability and annotation quality.

\noindent\textbf{Automated Pre-segmentation with Semantic Priors.}  
We begin by leveraging intrinsic modeling cues present in artist-created 3D assets, such as mesh connectivity, UV layout boundaries, and material assignments. To align these low-level cues with high-level semantics, we integrate a 3D-aware segmentation framework built upon SAM-2~\cite{kirillov2023segmentany} and Samesh~\cite{tang2024samesh}, enhanced with geometric priors specific to procedural 3D modeling workflows. This hybrid approach produces an initial over-segmented partition of each object, deliberately erring on the side of finer granularity. Over-segmentation is preferred because it provides annotators with atomic building blocks that can be reliably merged, whereas under-segmentation often leads to irreversible loss of part boundaries.

\noindent\textbf{Human-in-the-Loop Refinement.}  
All pre-segmented results undergo careful manual curation using a custom Blender-based annotation interface. Annotators first filter out unsuitable assets—such as those with excessive topological complexity, non-manifold geometry, or ambiguous part semantics. They then refine the segmentation by: (1) merging fragments that belong to the same functional or visual unit (e.g., the back and seat of a chair), and (2) splitting regions that conflate distinct components (e.g., separating armrests from chair legs). The refinement protocol emphasizes semantic clarity, structural symmetry, and compatibility with downstream generation tasks. Critically, textures and material properties are preserved throughout this process, ensuring that each extracted part remains visually coherent and renderable.

\noindent\textbf{Part-Aware Textual Captions.}  
To enable vision-language conditioning in part-level generation, we generate descriptive captions for every part. For each object-part pair, we render multi-view RGB images of both the full object and the isolated part. We then identify the view that maximizes the visible overlap between the part and its context within the whole object. A composite image is formed by overlaying a bounding box around the part in the full-object render. This image is fed to a state-of-the-art vision-language model (Qwen2.5-VL~\cite{wang2024qwen2} in our setting) to produce a natural language description that captures the part’s shape, appearance, material, and functional or spatial relationship to the parent object.

\textit{PartVerse-XL} not only provides an order-of-magnitude increase in scale over prior part datasets but also establishes a new standard for semantic consistency, texture fidelity, and multimodal alignment—making it uniquely suited for training and evaluating next-generation part-aware 3D generative models.

\end{document}